\documentclass[runningheads]{llncs}
\usepackage{graphicx,amsmath,amssymb}
\usepackage{algorithm}
\usepackage{algpseudocode}

\usepackage[caption=false]{subfig} 
\usepackage[usenames, dvipsnames]{color}
\newcommand{\bA}{ \mathbf{A} }

\newcommand{\bb}{ \mathbf{b} }

\newcommand{\bc}{ \mathbf{c} }

\newcommand{\bu}{ \mathbf{u} }

\newcommand{\bw}{ \mathbf{w} }

\newcommand{\bx}{ \mathbf{x} }

\newcommand{\balpha}{ \boldsymbol{\alpha} }

\newcommand{\bzero}{ \mathbf{0} }

\newcommand{\cinit}{\bc_{\text{ini}}}

\newcommand{\clearned}{\bc_{\text{lrn}}}

\newcommand{\xtrue}{\bx_{\text{tru}}}
\newcommand{\xlearned}{\bx_{\text{lrn}}}

\newcommand{\Ainit}{\bA_{\text{ini}}}

\newcommand{\Alearned}{\bA_{\text{lrn}}}

\newcommand{\binit}{\bb_{\text{ini}}}

\newcommand{\blearned}{\bb_{\text{lrn}}}

\newcommand{\winit}{{\bw}_{\text{ini}}}
\newcommand{\wtrue}{{\bw}_{\text{tru}}}
\newcommand{\wlearned}{{\bw}_{\text{lrn}}}

\begin{document}
\title{Deep Inverse Optimization
}
\author{Yingcong Tan\inst{1}
\and
Andrew Delong\inst{2}
\and
Daria Terekhov\inst{1}
}
\authorrunning{Y. Tan et al.}

\institute{Department of Mechanical, Industrial and Aerospace Engineering,\\ Concordia University \\
\email{t\_yingco@encs.concordia.ca\\ daria.terekhov@concordia.ca}
\and
\email{andrew.delong@gmail.com}
}

\maketitle              
\begin{abstract}
Given a set of observations generated by an optimization process, the goal of inverse optimization is to determine likely parameters of that process. We cast inverse optimization as a form of deep learning. Our method, called {\em deep inverse optimization}, is to unroll an iterative optimization process and then use backpropagation to learn parameters that generate the observations. We demonstrate that by backpropagating through the interior point algorithm we can learn the coefficients determining the cost vector and the constraints, independently or jointly, for both non-parametric and parametric linear programs, starting from one or multiple observations. With this approach, inverse optimization can leverage concepts and algorithms from deep learning.

\keywords{inverse optimization  \and deep learning \and interior point}
\end{abstract}
\section{Introduction}
The potential for synergy between optimization and machine learning is well-recognized~\cite{Bengio18}, with recent examples including~\cite{Bonami18,Fischetti18,Mahmood18b}. Our work uses machine learning for \emph{inverse} optimization. Consider a parametric linear optimization problem, PLP$(\bu, \bw)$: 
\begin{equation}\label{eq:linear_fop}
\begin{alignedat}{1}
\quad \underset{\bx}{\text{minimize}} & \quad \bc(\bu, \bw)' \bx\\
\text{subject to}            & \quad \bA(\bu, \bw)\bx \;\;\; \le \; \bb(\bu, \bw), 
\end{alignedat}
\end{equation}
where $\bx \in \mathbb{R}^{d}$ and $\bc(\bu, \bw) \in \mathbb{R}^{d}$, $\bA(\bu, \bw) \in \mathbb{R}^{d \times m}$ and $\bb(\bu, \bw) \in \mathbb{R}^{m}$ are all functions of features $\bu$ and weights $\bw$.  
Let $\xtrue^{n}$ be an optimal solution to PLP$(\bu^{n}, \wtrue)$. 
Given a set of observed optimal solutions, $\{\xtrue^{1}, \xtrue^{2}, \dots, \xtrue^{N}\}$, for observed conditions $\{\bu^{1}, \bu^{2}, \dots, \bu^{N}\}$, the goal of inverse optimization (IO) is to determine values of optimization process parameters $\bw$ that generated the observed optimal solutions. Applications of IO range from medicine (e.g., imputing the importance of treatment sub-objectives from clinically-approved radiotherapy plans \cite{Chan14}) to energy (e.g., predicting the behaviour of price-responsive customers \cite{Gallego17}). 

Fundamentally, IO problems are learning problems: each $\bu^n$ is a feature vector and $\xtrue^n$ is its corresponding target; the goal is to learn model parameters $\bw$ that minimize some loss function. In this paper, we cast inverse optimization as a form of deep learning. Our method, called {\em deep inverse optimization}, is to unroll an iterative optimization process and then use backpropagation to learn model parameters that generate the observations/targets.

\begin{figure}
\includegraphics[width=\textwidth]{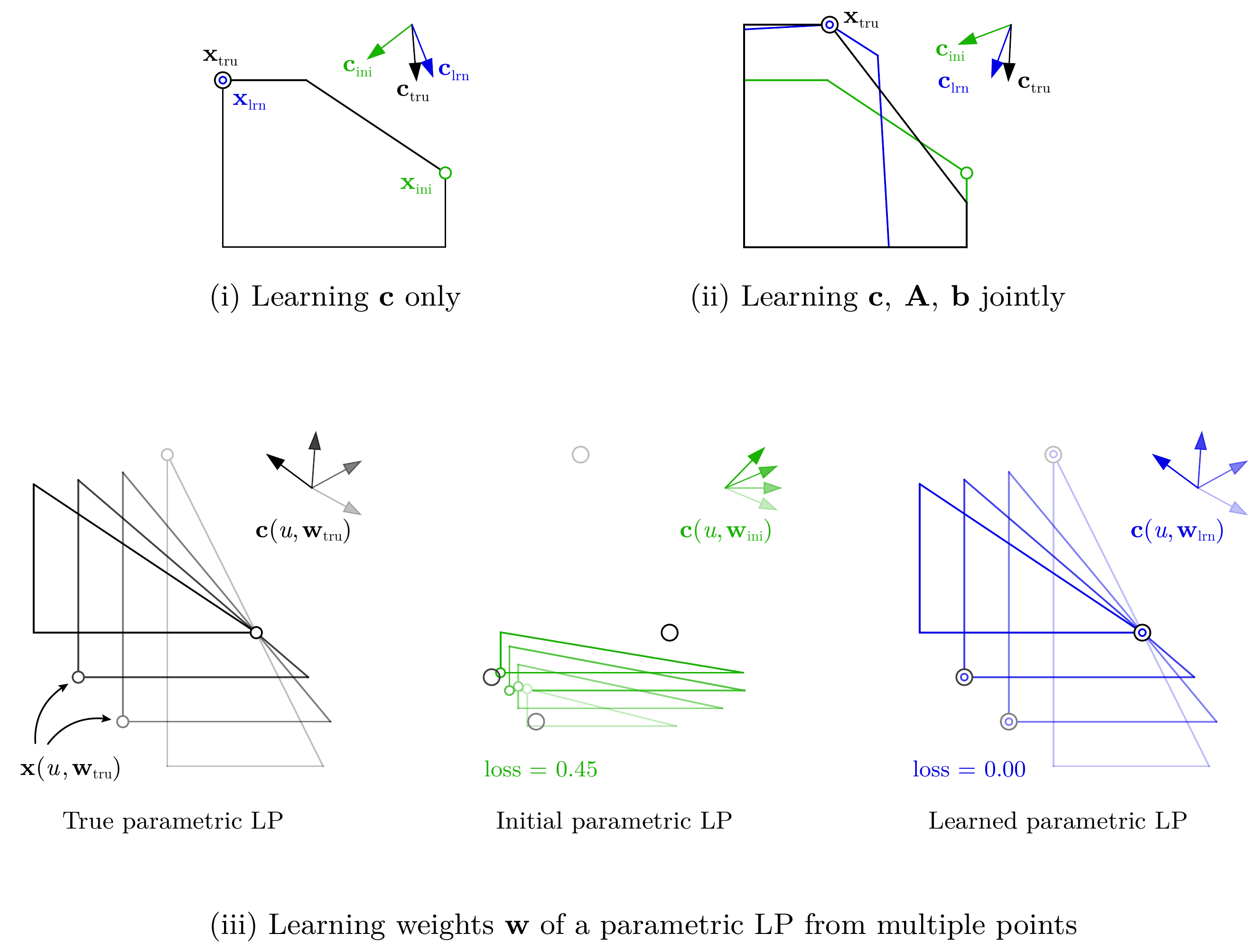}
\caption{Three IO learning tasks in non-parametric and parametric linear programs.} \label{fig:fig1}
\end{figure}

Figure \ref{fig:fig1} shows the actual result of applying our deep IO method to three inverse optimization learning tasks. 
The top panel illustrates the \emph{non-parametric}, single-point variant of model \eqref{eq:linear_fop} --- the case when exactly one $\xtrue$ is given --- a classical problem in IO (see  \cite{Ahuja01,Chan18}). 
In Figure~\ref{fig:fig1}~(i), only $\bc$ needs to be learned: starting from an initial cost vector $\cinit$, our method finds $\clearned$ which makes $\xtrue$ an optimal solution of the LP by minimizing $\|\xtrue - \xlearned\|^2$. 
In Figure~\ref{fig:fig1}~(ii), starting from $\cinit$, $\Ainit$ and $\binit$, our approach finds $\clearned$, $\Alearned$ and $\blearned$ which make $\xtrue$ an optimal solution of the learned LP through minimizing $\|\xtrue - \xlearned\|^2$.   
 
Figure~\ref{fig:fig1}~(iii) shows learning $\bw = [w_0, w_1]$ for the \emph{parametric} problem instance 
\begin{equation}\label{eq:example_plp}
\begin{alignedat}{1}
\quad \underset{\bx}{\text{minimize}} & \quad \cos(w_0 + w_1u)x_1 + \sin(w_0 + w_1u)x_2 \\
\text{subject to}            & \quad -x_1 \le 0.2 w_0 u, \\
& \quad -x_2 \le -0.2 w_1 u, \\
& \quad w_0x_1 + (1 + \textstyle \frac{1}{3} w_1 u)x_2 \le w_0 + 0.1 u.
\end{alignedat}
\end{equation}
Starting from $\winit=[0.2, 0.4]$ with a loss (mean squared error) of 0.45, our method is able to find $\wlearned=[1.0, 1.0]$ with a loss of zero, thereby making $\xtrue^n$ optimal solutions of \eqref{eq:example_plp} for $u$ values $\{-1.5, -0.5, 0.5, 1.5\}$.
Given newly observed $u$ values, in this example $\wlearned$ would predict correct decisions.
In other words, the learned model generalizes well.

The contributions of this paper are as follows. We propose a general framework for inverse optimization based on deep learning. This framework is applicable to learning coefficients of the objective function and constraints, individually or jointly; minimizing a general loss function; learning from a single or multiple observations; and solving both non-parametric and parametric problems. As a proof of concept, we demonstrate that our method obtains effectively zero loss on many randomly generated linear programs for all three types of learning tasks shown in Figure \ref{fig:fig1}, and always improves the loss significantly. 
Such a numerical study on randomly generated non-parameteric and parametric linear programs with multiple learnable parameters has not previously been published for any IO method in the literature. Finally, to the best of our knowledge, we are the first to use unrolling and backpropagation for constrained inverse optimization.

We explain how our approach differs from methods in inverse optimization and machine learning in Section \ref{sec:related_work}. We present our deep IO framework in Section \ref{sec:methodology} and our experimental results in Section \ref{sec:experiments}. Section \ref{sec:discussion} discusses both the generality and the limitations of our work, and Section \ref{sec:conclusion} concludes the paper. 

\section{Related Work}\label{sec:related_work}
The goal of our paper is to develop a general-purpose IO approach that is applicable to problems for which theoretical guarantees or efficient exact optimization approaches are difficult or impossible to develop. Naturally, such a general-purpose approach will not be the method of choice for all classes of IO problems. In particular, for non-parametric linear programs, closed-form solutions for learning the $\bc$ vector (Figure \ref{fig:fig1} (i)) and for learning the constraint coefficients have been derived by Chan et al.~\cite{Chan18,Chan18b} and Chan and Kaw~\cite{Chan18c}, respectively. However, learning objective and constraint coefficients jointly (Figure \ref{fig:fig1} (ii)) has, to date, received little attention. To the best of our knowledge, this task has been investigated only by Troutt et al.~\cite{Troutt08,Troutt05}, who referred to it as linear system identification, using a maximum likelihood approach. However, their approach was limited to two dimensions~\cite{Troutt05} or required the coefficients to be non-negative~\cite{Troutt08}. 

In the parametric optimization setting, Keshavarz et al.~\cite{Keshavarz11} develop an optimization model that encodes KKT optimality conditions for imputing objective function coefficients of a convex optimization problem. Aswani et al.~\cite{Aswani18} focus on the same problem under the assumption of noisy measurements, developing a bilevel problem and two algorithms which are shown to maintain statistical consistency. Saez-Gallego and Morales~\cite{Gallego17} address the case of learning $\bc$ and $\bb$ jointly in a parametric setting where the $\bb$ vector is assumed to be an affine function of a regressor. The general case of learning the weights of a parametric linear optimization problem \eqref{eq:linear_fop} where $\bc$, $\bA$ and $\bb$ are functions of $\bu$ (Figure \ref{fig:fig1} (iii)) has not been addressed in the literature.

Recent work in machine learning~\cite{Barmann18,Barmann17,Dong18b} views inverse optimization through the lens of online learning, where new observations appear over time rather than as one batch. Our approach may be applicable in online settings, but we focus on generality in the batch setting and do not investigate real-time cases.  

Methodologically, our unrolling strategy is  similar to McLaurin et al.~\cite{Maclaurin15} who directly optimize the hyperparameters of a neural network training procedure with gradient descent. Conceptually, the closest papers to our work are by Amos and Kolter~\cite{Amos2017} and Donti, Amos and Kolter~\cite{Donti2017}. However, these papers are written independently of the inverse optimization literature. 
Amos and Kolter~\cite{Amos2017} present the OptNet framework, which integrates a quadratic optimization layer in a deep neural network. The gradients for updating the coefficients of the optimization problem are derived through implicit differentiation. 
This approach involves taking matrix differentials of the KKT conditions for the optimization problem in question, while our strategy is based on allowing a deep learning framework to unroll an existing optimization procedure. Their method has efficiency advantages, while 
our unrolling approach is easily applicable, including to processes for which the KKT conditions may not hold or are difficult to implicitly differentiate. We include a more in-depth discussion in Section~\ref{sec:discussion}. 

\section{Deep Learning Framework for Inverse Optimization}
\label{sec:methodology}

The problems studied in inverse optimization are learning problems: given features $\bu^n$ and corresponding targets $\xtrue^n$, the goal is to learn parameters of a forward optimization model that generate $\xtrue^n$ as its optimal solutions. A complementary view is that inverse optimization is a learning technique specialized to the case when the observed data is coming from an optimization process. Given this perspective on inverse optimization and motivated by the success of deep learning for a variety of learning tasks in recent years (see~\cite{lecun2015deep}), this paper develops a deep learning framework for inverse optimization problems. 

Deep learning is a set techniques for training the parameters of a sequence of transformations (layers) chained together.
The more intermediate layers, the `deeper' the architecture. We refer the reader to the textbook by Goodfellow, Bengio and Courville~\cite{Goodfellow16} for additional details about deep learning.  
The features of the intermediate layers can be trained/learned through backpropagation, an automatic differentiation technique that computes the gradient of an output with respect to its input through the layers of a neural network, starting from the final layer all the way to the initial one. This method efficiently computes an update to the weights of the model~\cite{Rumelhart86}. Importantly, current machine learning libraries such as PyTorch provide built-in backpropagation capabilities~\cite{Paszke17} that allow for wider use of deep learning. 
Thus, our \emph{deep inverse optimization} framework iterates between solving the forward optimization problem using an iterative optimization algorithm and backpropagating through the steps (layers) of that algorithm to improve the estimates of learnable  parameters (weights) of the forward process.  

\begin{algorithm}
\caption{Deep inverse optimization framework.}
\label{algo:approach}
\begin{algorithmic}[1]
\Statex Input: $\winit$; $(\bu^n, \xtrue^n)$ for $n = 1, \;..\; N$, 
\Statex Output: $\wlearned$ 
\State $\bw \leftarrow \winit$ 
\For{$s$ in $1 \; ..$ \texttt{max$\_$steps}}
\State $\Delta\bw \leftarrow \bzero$
  \For{$n$ in $1 \; .. \; N$}
        \State $\bx \leftarrow \textbf{FO}(\bu^n, \bw)$ \Comment{Solve forward problem}
        \State $\ell \leftarrow \mathcal{L}(\bx, \xtrue^n)$ \Comment{Compute loss}
        \State $\Delta\bw \; \leftarrow \Delta\bw +  \frac{\partial\ell}{\partial \bw}$ \Comment{Accumulate gradient by backprop}
      \EndFor
      \State $\beta \leftarrow \texttt{line\_search}(\bw, \balpha\cdot\frac{\Delta\bw}{N})$ \Comment{Find safe step size}
      \State $\bw \leftarrow \bw - \beta \balpha\cdot\frac{\Delta\bw}{N}$ \Comment{Update weights}
\EndFor
\State Return $\bw$		
\end{algorithmic}
\end{algorithm}

Our approach, shown in Algorithm \ref{algo:approach}, takes the pairs $(\bu^{n}, \xtrue^{n})$, $n = 1, \; .., \; N$, as input, and starts by initializing $\bw = \winit$. For each $n$, the forward optimization problem (\textbf{FO})
is solved with the current weights (line 5), and the loss between the resulting optimal solution $\bx$ and $\xtrue$ is computed (line 6). The gradient of the loss function with respect to $\bw$ is computed by backpropagation through the layers of the forward process. In line 9, line search is used to determine the step size, $\beta$, for updating the weights: $\beta$ is reduced by half if infeasibility or unboundedness is encountered until a value is found that will lead to loss reduction or $\beta < 10^{-8}$, in which case early algorithm termination is triggered.
Finally, in line 10, the weights are updated using the average gradient, step size $\beta$, and $\balpha$, a vector representing the component-wise learning rates for $\bw$. 

\begin{figure}
\includegraphics[width=\textwidth]{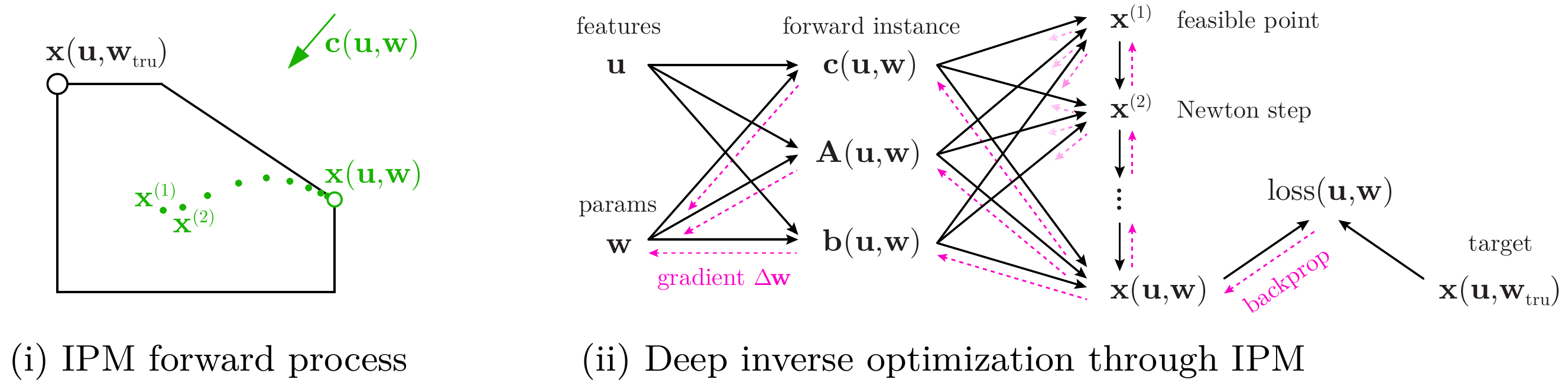}
\caption{Deep inverse optimization framework.} \label{fig:fig2}
\end{figure}

Importantly, our framework is applicable in the context of any differentiable, iterative forward optimization procedure. In principle, parameter gradients are automatically computable even with non-linear constraints or non-linear objectives, so long as they can be expressed through standard differentiable primitives. Our particular implementation uses the barrier interior point method (IPM) as described by Boyd and Vandenberghe~\cite{Boyd04}, as our forward optimization solver. The IPM forward process is illustrated in Figure \ref{fig:fig2} (i): the central path taken by IPM is illustrated for the current $\bu$ and $\bw$, which define both the current feasible region and the current $\bc(\bu,\bw)$. As shown in Figure \ref{fig:fig2} (ii), backpropagation starts from the computation of the loss function between a (near) optimal forward optimization solution $\bx(\bu, \bw)$ and the target $\bx(\bu, \wtrue)$ and proceeds backward through all the steps of IPM, i.e., $\bx(\bu, \bw)$ to $\bx^{(1)}$, the starting point of IPM, to the forward instance parameters and finally $\bw$ to compute $\Delta \bw$.
In practice, backpropagating all the way to $\bx^{(1)}$ may not be necessary for computing accurate gradients; see Section~\ref{sec:discussion}.

The framework requires setting three main hyperparameters: $\winit$, the initial weight vector; $\texttt{max\_steps}$, the total number of steps allotted to the training; and $\balpha$, the learning rates for the different components of $\bw$. The number of additional hyperparameters depends on the forward optimization process.

\section{Experimental Results} \label{sec:experiments}
In this section, we demonstrate the application of our framework on randomly-generated LPs for the three types of problems shown in Figure \ref{fig:fig1}: learning $\bc$ in the non-parametric case; learning $\bc$, $\bA$ and $\bb$ together in the non-parametric case; and learning $\bw$ in the parametric case.

\subsubsection{Implementation} Our framework is implemented in Python, using PyTorch version 0.4.1 and its built-in backpropagation capabilities~\cite{Paszke17}.
All numerical operations are carried out with PyTorch tensors and standard PyTorch primitives, including the matrix inversion at the heart of the Newton step. 

\subsubsection{Hyperparameters} 
We limit learning to $\texttt{max\_steps}=200$ in all experiments. Four additional hyperparameters are set in each experiment:
\begin{itemize}
\item $\epsilon$, which controls the precision and termination of IPM;
\item $t^{(0)}$: the initial value of the barrier IPM sharpness parameter $t$; 
\item $\mu$: the factor by which $t$ is increased along the IPM central path;
\item $\balpha$: the vector of per-parameter learning rates, which in some experiments is broken down into $\balpha_{\bc}$ and $\balpha_{\bA\bb}$. 

\end{itemize}
In all experiments, the $\epsilon$ hyperparameter is either a constant $10^{-5}$ or decays exponentially from $0.1$ to $10^{-5}$ during learning. The decay is a form of graduated optimization~\cite{Blake87}, and tends to help performance when using the MSE loss.

\subsubsection{Baseline LPs} To generate problem instances, we first create a set of baseline LPs with $d$ variables and $m$ constraints by sampling at least $d$ random points from $\mathcal{N}(0, 1)$, and then construct the convex hull via the  \emph{scipy.spatial.convexhull} package~\cite{convexHull}. We generate 50 LP instances for each of the following six problem sizes: $d = 2$ and $m \in \{4, 8, 16\}$ and $d = 10$, $m \in  \{20, 36, 80\}$. 
Our experiments focus on inequality constraints.
We observed that our method can work for equality constrained instances, but we did not systematically evaluate equality constraints and we leave that for future work.

\begin{figure}
\includegraphics[width=\textwidth]{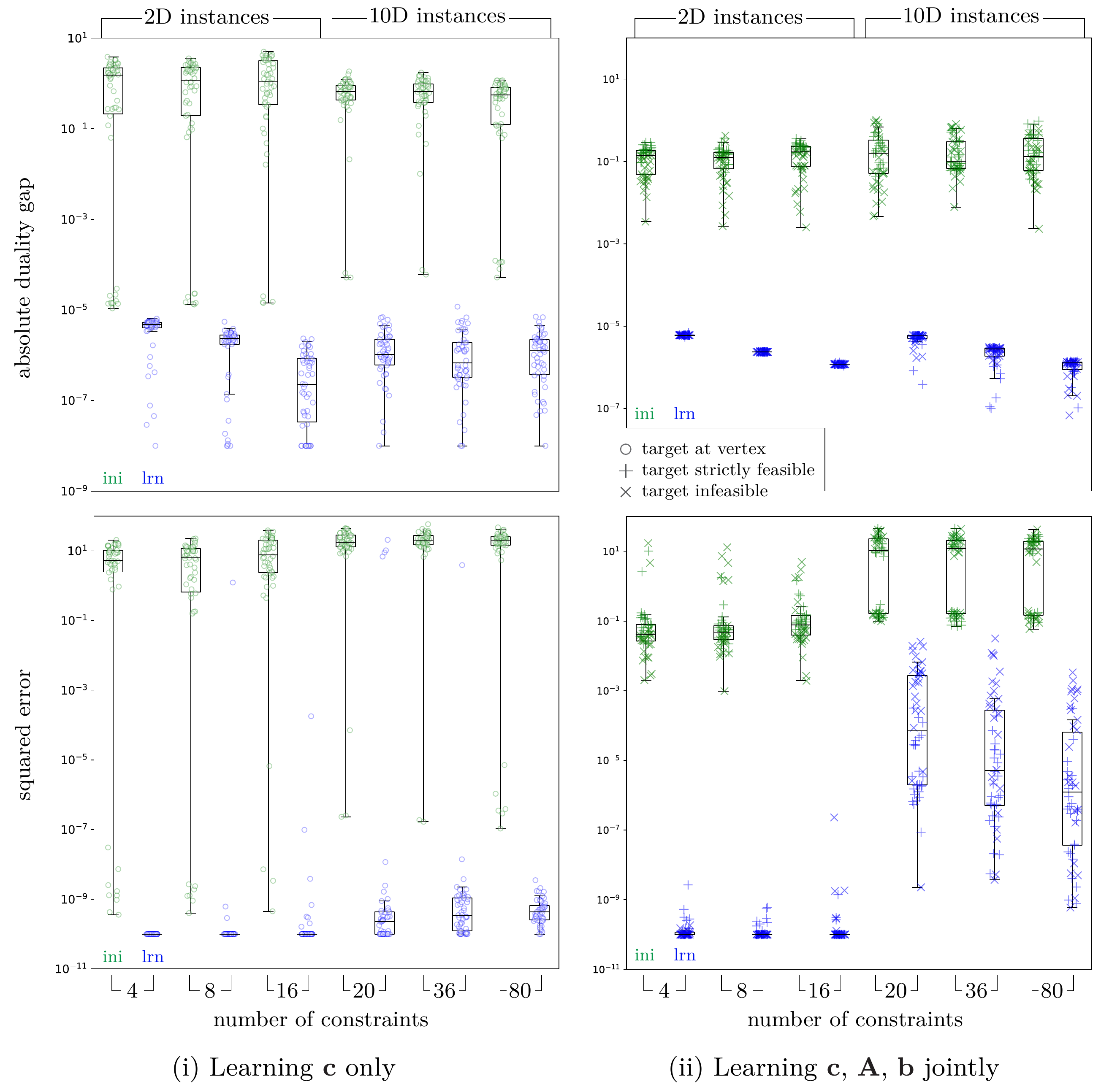}
\caption{Learning in non-parametric IO problems.} \label{fig:fig3}
\end{figure}

\subsection{Non-Parametric}
We first demonstrate the performance of our method for learning $\bc$ only, and learning $\bc$, $\bA$ and $\bb$ jointly, on the single-point variant of model \eqref{eq:linear_fop}, i.e., when a single optimal target $\xtrue$ is given, a classical problem in IO~\cite{Ahuja01}. 
We use two loss functions, absolute duality gap (ADG) and squared error (SE), defined as follows:
\begin{eqnarray}
\text{ADG} &=& \clearned'|\xtrue - \xlearned|, \\
\text{SE} &=& \|\xtrue - \xlearned\|^2_2,
\end{eqnarray}
the first of which is a classical performance metric in IO~\cite{Chan14} and the second is a standard metric in machine learning.

\subsubsection{Learning $\bc$ only} 
To complete instance generation for this experiment, we randomly select one vertex of the convex hull to be $\xtrue$ for each of the 50 baseline LP instances and for each of the six $(m, d)$ combinations. 

Initialization is done by sampling each parameter of $\cinit$ from $\mathcal{N}(0,1)$. 
We implement a randomized grid search by sampling 20 random combinations of the following three hyperparameter sets: $t^{(0)} \in \{0.5, 1, 5, 10\}$, $\mu \in \{1.5, 2, 5, 10,20\}$, and 
$\balpha_{\bc} \in \{1, 10, 100, 1000\}$. As in other applications of deep learning, it is not clear which hyperparameters will work best for a particular problem instance. For each instance we run our algorithm with the same 20 hyperparameter combinations, reporting the best final error values. 

Figure \ref{fig:fig3}~(i) shows the results of this experiment for ADG and SE loss. In both cases, our method is able to reliably learn $\bc$: in fact, for all instances, the final error is under $10^{-4}$, while the majority of initial errors are above $10^{-1}$. There is no clear pattern in the performance of the method as $m$ and $d$ change for ADG; for SE, the final loss is slightly bigger for higher $d$.

\subsubsection{Learning $\bc$, $\bA$, $\bb$ jointly} 
Our approach to instance generation here is to start with each baseline LP and generate a strictly feasible or infeasible target within some reasonable proximity of an existing vertex. The algorithm is then forced to learn a new $\bc, \bA, \bb$ that generate the target, which is not an optimum for the initial LP. To make this task more challenging, we also perturb $\bc$ so that it is not initialized too close to the optimal direction.

For each of the 50 baseline LP feasible regions, we generate a $\bc \sim \mathcal{N}(0,1)$ and compute its optimal solution $\bx^*$. To generate an infeasible target we set $\xtrue = \bx^* + \eta$ where $\eta \sim U[-0.2, 0.2]$. We similarly generate a challenging $\cinit$ by corrupting $\bc$ with noise from $U[-0.2, 0.2]$. 
To generate a strictly feasible target near $\bx^*$, we set $\xtrue = 0.9 \bx^* + 0.1 \bx'$ where $\bx'$ is a uniformly random point within the feasible region generated by Dirichlet-weighted combination of all vertices; this method was used  because adding noise in 10 dimensions almost always results in an infeasible target. 

In summary, we generate new LP instances with the same feasible region as the baseline LPs but a corrupted $\cinit$ and one feasible and one infeasible target. The goal is to demonstrate the ability of our algorithm to detect the change and also move the constraints and objective so that the feasible/infeasible target becomes a vertex optimum. For each of the six problem sizes, we randomly split the 50 instances into two subsets, one with feasible and the other with infeasible targets. For ADG loss we set $\epsilon = 10^{-5}$ and for SE we use the $\epsilon$ decay strategy. In practice, this decay strategy is similar to putting emphasis on learning $\bc$ in the initial iterations and ending with emphasis on constraint learning.

The values of hyperparameters $\balpha_{\bc} $ and $\balpha_{\bA \bb}$ are independently selected from $\{0.1, 1, 10\}$ and concatenated into one learning rate vector $\balpha$. We generate 20 different hyperparameter combinations. We run our algorithm on each instance with all hyperparameter combinations and record the value of the best trial. 

Figure \ref{fig:fig3}~(ii) shows the results of this experiment for ADG and SE loss. In both cases, our method is able to learn model parameters that result in median loss of under $10^{-4}$. For ADG, our method performs equally well for all problem sizes, and there is not much difference in the final loss for feasible and infeasible targets. For SE, however, the final loss is larger for higher $d$ but decreases as $m$ increases. Furthermore, there is a visible difference in performance of the method on feasible and infeasible points for 10-dimensional instances: learning from infeasible targets becomes a more difficult task.

\subsection{Parametric}
Several aspects of the experiment for parametric LPs are different from the non-parametric case. First, we train by minimizing $\text{MSE}(\bw)$, defined as 
\begin{eqnarray}
\text{MSE}(\bw) &=& \frac{1}{N} \sum_{n=1}^{N}\|\bx(\bu^n, \wtrue) - \bx(\bu^n, \bw)\|^2_2. \label{eq:MSE}
\end{eqnarray}
We chose the mean of SE loss instead of the mean of ADG loss for the parametric experiments because it is only zero if the targets are all feasible, which is not necessarily required for ADG to be zero. This makes the SE loss more difficult from a learning point of view, but also leads to more intuitive notion of success. See Section~\ref{sec:discussion} for discussion. In the parametric case, we also assess how well the learned PLP generalizes, by evaluating its MSE$(\wlearned)$ on a held-out test set.  

To generate parametric problem instances, we again started from the baseline LP feasible regions. To generate a true PLP, we used six weights to define linear functions of $u$ for all elements of $\bc$, all elements of $\bb$, and one random element in each row of $\bA$. For example, for 2-dimensional problems with four constraints, our instances have the following form:

\begin{equation}\label{eq:linear_PLP}
\begin{alignedat}{1}
\quad \underset{\bx}{\text{minimize}} & \quad  (c_1+w_1+w_2u)x_1 + (c_2+w_1+w_2u)x_2 \\
\text{subject to}  & \quad 
\begin{bmatrix}
    a_{11}\phantom{{}+w_3+w_4u}      & a_{12}           +w_3+w_4u \\
    a_{21}\phantom{{}+w_3+w_4u}      & a_{22}           +w_3+w_4u \\
    \;\;a_{31}           +w_3+w_4u   \;\;& a_{32}\phantom{{}+w_3+w_4u} \\
    a_{41}\phantom{{}+w_3+w_4u}      & a_{42}           +w_3+w_4u \\
\end{bmatrix}
\le
\begin{bmatrix}
    b_{1}+w_5+w_6u \\
    b_{2}+w_5+w_6u \\
    b_{3}+w_5+w_6u \\
    b_{4}+w_5+w_6u 
\end{bmatrix}
\end{alignedat}
\end{equation}
Specifically, the ``true PLP'' instances are generated by setting $w_1,w_3,w_5 = 0$ and $w_2,w_4,w_6 \sim  \mathcal{N}(0, 0.2)$. This ensures that when $u=0$ the feasible region of the true PLP matches the baseline LP.
For each true PLP, we find a range $[u_{min},u_{max}] \subseteq [-1, 1]$ over which the resulting PLP remains bounded and feasible. To find this `safe' range we evaluate $u$ at increasingly large values and try to solve the corresponding LP, expanding $[u_{min},u_{max}]$ if successful.
For each true PLP, we generate 20 equally spaced training points spanning $[u_{min},u_{max}]$.
We also sample 20 test points $u$ sampled uniformly from $[u_{min},u_{max}]$. 
We then initialize learning from a corrupted PLP by setting $\winit = \wtrue + \mathbf{\eta}$ where each element of $\eta \sim U[-0.2, 0.2]$.

Hyperparameters are sampled as $t^{(0)} \in \{0.5, 1, 5, 10\}$, $\mu \in \{1.5, 2, 5, 10, 20\}$ and  $\balpha_{\bA \bb} \in \{1, 10\}$, and $\balpha_{\bc}$ is then chosen to be a factor of $\{0.01, 1, 100\}$ times $\balpha_{\bA \bb}$, i.e., a relative learning rate. Here, $\balpha_{\bc}$ and $\balpha_{\bA \bb}$ control the learning rate of parameters within $\bw$ that determine $\bc$ and $(\bA, \bb)$, respectively. In total, we generate 20 different hyperparameter combinations. We run our algorithm on each instance with all hyperparameter combinations and record the best final error value. 
A constant value of $\epsilon = 10^{-5}$ is used.

%**figure 4 ******************************************
\begin{figure}
\includegraphics[width=\textwidth]{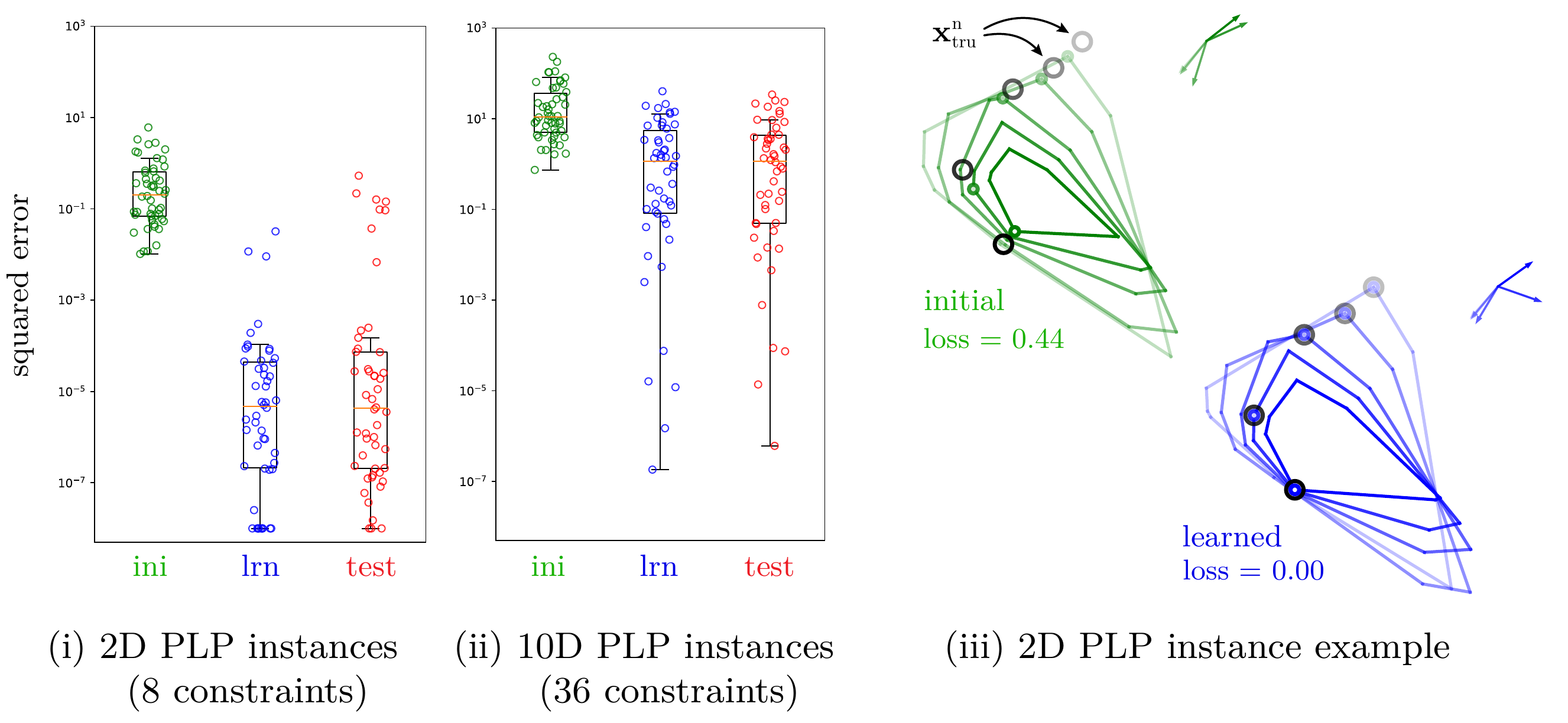}
\caption{Learning in non-parametric IO problems.} \label{fig:fig4}
\end{figure}
%**figure 4 ******************************************

We demonstrate the performance of our method on learning parametric LPs of the form shown in \eqref{eq:linear_PLP} with $d = 2 $, $m = 8$, and $d = 10$, $m = 36$. In Figure \ref{fig:fig4}, we report two metrics evaluated on the training set, namely MSE$(\winit)$ and MSE$(\wlearned)$, and one metric for the test set, MSE$(\wlearned)$. Figure \ref{fig:fig4} (iii) shows an example of an instance with $d = 2$, $m = 8$ from the training set. We see that, overall, our deep learning method works well on 2-dimensional problems with the training and testing error both being much smaller than the initial error. In the vast majority of cases the test error is also comparable to training error, though there are a few cases where it is worse, which indicates a failure to generalize well. For 10D instances, the algorithm significantly improves MSE$(\wlearned)$ over the initialization MSE$(\winit)$, but in most cases fails to drive the loss to zero, either due to local minima or slow convergence. Again, performance on the test set is similar to that on training set. 

\section{Discussion}\label{sec:discussion}

The conceptual message that we wish to reinforce is that inverse optimization should be viewed as a form of deep learning,
and that unrolling gives easy access to the gradients of any parameter used directly or indirectly in the forward optimization process.
There are many aspects to this view that merit further exploration.
What kind of forward optimization processes can be inversely optimized this way?
Which ideas and algorithms from the deep learning community will help?
Are there aspects of IO that make gradient-based learning more challenging than in deep learning at large?
Conclusive answers are beyond the scope of this paper, but we discuss these and other questions below.

\vspace{0.9em}
\noindent {\bf Generality and applicability.}\;
As a proof of concept, this paper uses linear programming for the forward problems and IPM with barrier method as the forward optimization process.
In principle, the framework is applicable to any forward process for which automatic differentiation can be applied.
This observation does not mean that ours is the best approach for a specialized IO problem, such as learning $\bc$ from a single point~\cite{Chan18} or multiple points within the same feasible region~\cite{Chan18b}, but it provides a new strategy.

The practical message of our paper is that, when faced with novel classes or novel parameterizations of IO problems, the unrolling strategy provides convenient access to a suite of general-purpose gradient-based algorithms for solving the IO problem at hand.
This strategy is made especially easy by deep learning libraries that support dynamic `computation graphs' such as PyTorch.
Researchers working within this framework can rapidly apply IO to many differentiable forward optimization processes, without having to derive the algorithm for each case.
Automatic differentiation and backpropagation have enabled a new level of productivity for deep learning research, and they may do the same for inverse optimization research.
Applying deep inverse optimization does not require expertise in deep learning itself.

We chose IPM as a forward process because the inner Newton step is differentiable and because we expected the gradient to temperature parameter $t$ to have a stabilizing effect on the gradient.
For non-differentiable optimization processes, it may still be possible to develop differentiable versions.
In deep learning, many advances have been made by developing differentiable versions of traditionally discrete operations, such as memory addressing~\cite{graves2016hybrid} or sampling from a discrete distribution~\cite{maddison2016concrete}.
We believe the scope of differentiable forward optimization processes may similarly be expanded over time.

\vspace{0.9em}
\noindent {\bf Limitations and possible improvements.}\;
Deep IO inherits the limitations of most gradient-based methods.
If learning is initialized to the right ``basin of attraction'', it can proceed to a global optimum.
Even then, the choice of learning algorithm may be crucial.
When implemented within a steepest descent framework, as we have here, the learning procedure can get trapped in local minima or exhibit very slow convergence.
Such effects are why most instances in Figure~\ref{fig:fig4} (ii) failed to achieve zero loss.

In deep learning with neural networks, poor local minima become exponentially rare as the dimension of the learning problem increases~\cite{dauphin2014identifying,soudry2017exponentially}.
A typical strategy for training neural networks is therefore to over-parameterize (use a high search dimension) and then use regularization to avoid over-fitting to the data.
In deep IO, natural parameterizations of the forward process may not permit an increase in dimension, or there may not be enough observations for regularization to compensate, so local minima remain a potential obstacle.
We believe training and regularization methods specialized to low-dimensional learning problems such as by Sahoo et al.~\cite{sahoo2018learning} may be applicable here.

We expect that other techniques from deep learning, and from gradient-based optimization in general, will translate to deep IO.
For example, optimization techniques with second-order aspects such as momentum~\cite{sutskever2013importance} and L-BFGS~\cite{byrd1995limited} are readily available in deep learning frameworks.
Other deep learning `tricks' may be applicable to stabilizing deep IO.
For example, we observe that, when $\bc$ is normal to a constraint, the gradient with respect to $\bc$ can suddenly grow very large.
We stabilized this behaviour with line search, but a similar `exploding gradient' phenomenon exists when training deep recurrent neural networks, and gradient clipping~\cite{pascanu2012understanding} is a popular way to stabilize training.
A detailed investigation of applicable deep learning techniques is outside the scope of this paper.

Deep IO may be more successful when the loss with respect to the forward process can be annealed or `smoothed' in a manner akin to graduated non-convexity~\cite{Blake87}. Our $\epsilon$-decay strategy is an example of this, as discussed below.

Finally, it may be possible to develop hybrid approaches, combining gradient-based learning with closed-form solutions or combinatorial algorithms.

\vspace{0.9em}
\noindent {\bf Loss function and metric of success.}\;
One advantage of the deep inverse optimization approach is that it is can accommodate various loss functions, or combinations of loss functions, without special development or analysis.
For example one could substitute other $p$-norms, or losses that are robust to outliers, and the gradient will be automatically available.
This flexibility may be valuable.
Special loss functions have been important in machine learning, especially for structured output problems~\cite{hazan2010direct}.
The decision variables of optimization processes are likewise a form of structured output.

In this study we chose two classical loss functions: absolute duality gap and squared error.
The behaviour of our algorithm varied depending on the loss function used.
Looking at Figure~\ref{fig:fig3} (ii) it appears that deep IO performs better with ADG loss than with SE loss when learning $\bc, \bA, \bb$ jointly.
However, this performance is due to the theoretical property that ADG can be zero even when the observed target point is arbitrarily infeasible \cite{Chan18}. With ADG, all the IO solver needs to do is adjust $\bc, \bA, \bb$ so that $\xlearned - \xtrue$ is orthogonal to $\bc$, which in no way requires the learned model to be capable of generating $\xtrue$ as an optimum.
In other words, ADG is meaningful mainly when the true feasible region is known, as in Figure~\ref{fig:fig3} (i).
When the true region is unknown, SE prioritizes solutions that directly generate the observations $\xtrue^n$,
and may therefore be a more meaningful loss function.
That is why we used it for our parametric experiments depicted in Figure~\ref{fig:fig4}.

\begin{figure}
    \centering
    \includegraphics[width=\textwidth]{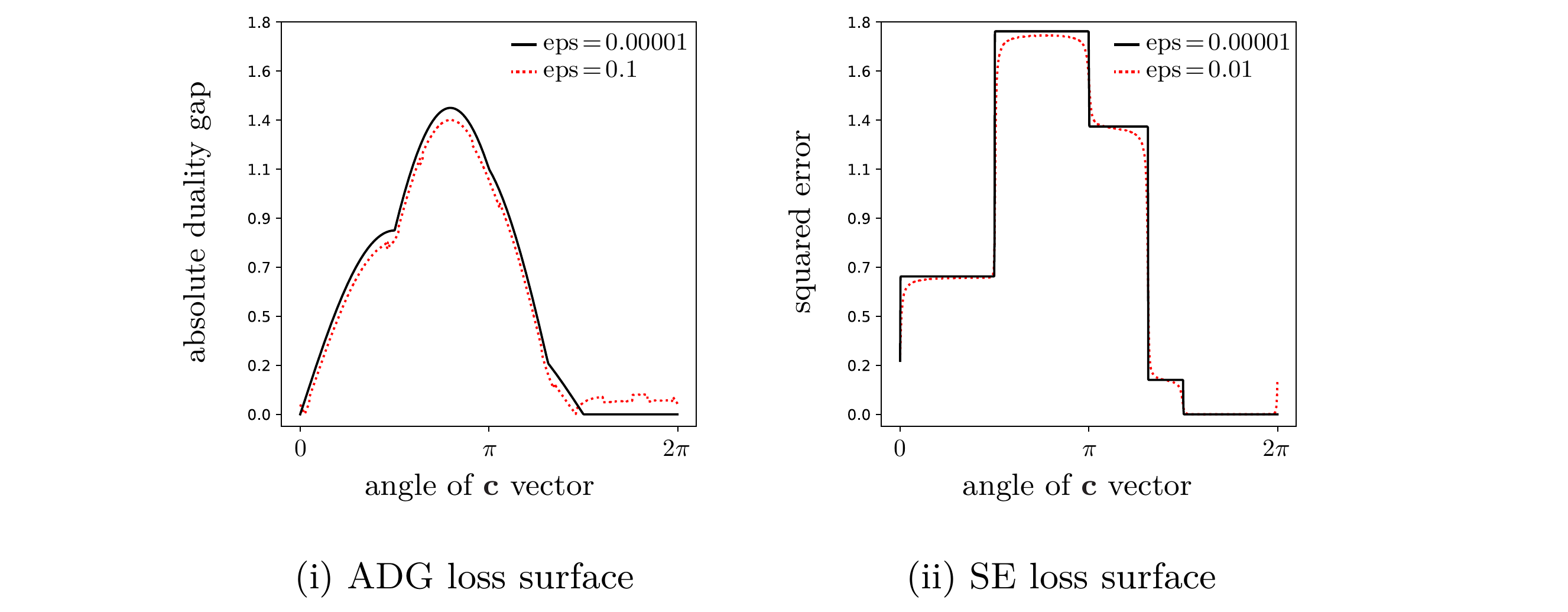}
    \caption{Loss surfaces for the feasible region and target shown in Figure 1 (i).}
    \label{fig:ADG_MSE_loss_surface}
\end{figure}

Minimizing the SE loss also appears to be more challenging for steepest descent.
To get a sense for the characteristics of ADG versus SE from the point of view of varying $\bc$, consider Figure~\ref{fig:ADG_MSE_loss_surface}, which depicts the loss for the IO problem in Figure~\ref{fig:fig1} (i) using both high precision ($\epsilon=10^{-5})$ and low precision ($\epsilon=0.1, 0.01$) for IPM.
Because the ADG loss is directly dependent on $\bc$, the loss varies smoothly even as the corresponding optimum $\bx^*$ stays fixed.
The SE loss, in contrast, is piece-wise constant; an instantaneous perturbation of $\bc$ will almost never change the SE loss in the limit of $\epsilon \rightarrow 0$.
Note that the gradients derived by implicit differentiation \cite{Amos2017} indicate $\frac{\partial \ell}{\partial \bc} = \mathbf{0}$ everywhere in the linear case, which would mean $\bc$ cannot be learned by gradient descent.
IPM can learn $\bc$ nonetheless because the barrier sharpness parameter $t$ smooths the loss, especially at low values.
The precision parameter $\epsilon$ limits the maximal sharpness during forward optimization, and so the gradient $\frac{\partial \ell}{\partial \bc}$ is not zero in practice, especially when $\epsilon$ is weak.
Notice that the SE loss surface becomes qualitatively smoother, whereas ADG is not fundamentally changed.
Also notice that when $\bc$ is normal to a constraint (when the optimal point is about to transition from one point to another) the gradient $\frac{\partial \ell}{\partial \bc}$ explodes even when the problem is smoothed.

\vspace{0.9em}
\noindent {\bf Computational efficiency.}\;
Our paper is conceptual and focuses on flexibility and the likelihood of success, rather than computational efficiency.
Many applications of IO are not real-time, and so we expect methods with running times on the order of seconds or minutes to be of practical use.
Still, we believe the framework can be both flexible and fast.

Deep learning frameworks are GPU accelerated and scale well with the size of an individual forward problem, so large instances are not a concern.
A bigger issue for GPUs is solving many small or moderate instances efficiently.
Amos and Kolter~\cite{Amos2017} developed a batch-mode GPU forward solver to address this.

What is more concerning for the unrolling strategy is that forward optimization processes can be very deep, with hundreds or thousands of iterations.
Backpropagation requires keeping all the intermediate values of the forward pass resident in memory, for later use in the backward pass.
The computational cost of backpropagation is comparable to that of the forward process, so there is no asymptotic advantage to skipping the backwards pass.
Although memory usage was small in our instances, if the memory usage is linear with depth, then at some depth the unrolling strategy will cease to be practical compared to Amos and Kolter's \cite{Amos2017} implicit differentiation approach.
However, we observed that for IPM most of the gradient contribution comes from the final ten  Newton steps before termination.
In other words, there is a vanishing gradient with depth, which means the gradient can be well-approximated in practice with {\em truncated backpropagation through time} (see~\cite{sutskever2013training} for review), which uses a small constant pool of memory regardless of depth.

In practice, we suggest that the unrolling approach is convenient during the development and exploration phase of IO research.
Once an IO model is proven to work, it can potentially be made more efficient by deriving the implicit gradients~\cite{Amos2017} and comparing them to the unrolled implementation as a reference.
Still, more important than improving any of these constants is to use asymptotically faster learning algorithms actively being developed in the deep learning community.

\section{Conclusion}\label{sec:conclusion}
We developed a deep learning framework for inverse optimization based on backpropagation through an iterative forward optimization process.
We illustrate the potential of this framework via an  implementation where the forward process is the interior point barrier method. Our results on linear non-parametric and parametric problems show promising performance. To the best of our knowledge, this paper is the first to explicitly connect deep learning and inverse optimization. 

\bibliographystyle{splncs04}
\bibliography{deep_inverse}

\begin{thebibliography}{10}
\providecommand{\url}[1]{\texttt{#1}}
\providecommand{\urlprefix}{URL }
\providecommand{\doi}[1]{https://doi.org/#1}

\bibitem{Ahuja01}
Ahuja, R.K., Orlin, J.B.: Inverse optimization. Operations Research
  \textbf{49}(5),  771--783 (2001)

\bibitem{Amos2017}
Amos, B., Kolter, J.Z.: {OptNet}: Differentiable optimization as a layer in
  neural networks. In: Proceedings of the 34th International Conference on
  Machine Learning, PMLR 70. pp. 136--145 (2017)

\bibitem{Aswani18}
Aswani, A., Shen, Z.J., Siddiq, A.: Inverse optimization with noisy data.
  Operations Research  \textbf{63}(3) (2018)

\bibitem{Barmann18}
B{\"a}rmann, A., Martin, A., Pokutta, S., Schneider, O.: An online-learning
  approach to inverse optimization. arXiv preprint arXiv:1810.12997  (2018)

\bibitem{Barmann17}
B{\"a}rmann, A., Pokutta, S., Schneider, O.: Emulating the expert: Inverse
  optimization through online learning. In: International Conference on Machine
  Learning. pp. 400--410 (2017)

\bibitem{Bengio18}
Bengio, Y., Lodi, A., Prouvost, A.: Machine learning for combinatorial
  optimization: a methodological tour d'horizon. arXiv preprint
  arXiv:1811.06128  (2018)

\bibitem{Blake87}
Blake, A., Zisserman, A.: Visual Reconstruction. MIT Press, Cambridge, MA, USA
  (1987)

\bibitem{Bonami18}
Bonami, P., Lodi, A., Zarpellon, G.: Learning a classification of mixed-integer
  quadratic programming problems. In: International Conference on the
  Integration of Constraint Programming, Artificial Intelligence, and
  Operations Research. pp. 595--604. Springer (2018)

\bibitem{Boyd04}
Boyd, S., Vandenberghe, L.: Convex optimization. Cambridge university press
  (2004)

\bibitem{byrd1995limited}
Byrd, R.H., Lu, P., Nocedal, J., Zhu, C.: A limited memory algorithm for bound
  constrained optimization. SIAM Journal on Scientific Computing
  \textbf{16}(5),  1190--1208 (1995)

\bibitem{Chan14}
Chan, T.C.Y., Craig, T., Lee, T., Sharpe, M.B.: Generalized inverse
  multi-objective optimization with application to cancer therapy. Operations
  Research  \textbf{62}(3),  680--695 (2014)

\bibitem{Chan18}
Chan, T.C.Y., Lee, T., Terekhov, D.: Goodness of fit in inverse optimization.
  Management Science  (2018)

\bibitem{Chan18c}
Chan, T.C.Y., Kaw, N.: Inverse optimization for the recovery of constraint
  parameters. arXiv preprint arXiv:1811.00726  (2018)

\bibitem{Chan18b}
Chan, T.C.Y., Lee, T., Mahmood, R., Terekhov, D.: Multiple observations and
  goodness of fit in generalized inverse optimization. arXiv preprint
  arXiv:1804.04576  (2018)

\bibitem{dauphin2014identifying}
Dauphin, Y.N., Pascanu, R., Gulcehre, C., Cho, K., Ganguli, S., Bengio, Y.:
  Identifying and attacking the saddle point problem in high-dimensional
  non-convex optimization. In: Advances in neural information processing
  systems. pp. 2933--2941 (2014)

\bibitem{Dong18b}
Dong, C., Chen, Y., Zeng, B.: Generalized inverse optimization through online
  learning. In: Advances in Neural Information Processing Systems. pp. 86--95
  (2018)

\bibitem{Donti2017}
Donti, P., Amos, B., Kolter, J.Z.: Task-based end-to-end model learning in
  stochastic optimization. In: Advances in Neural Information Processing
  Systems. pp. 5484--5494 (2017)

\bibitem{Fischetti18}
Fischetti, M., Jo, J.: Deep neural networks and mixed integer linear
  optimization. Constraints pp. 1--14 (2018)

\bibitem{Goodfellow16}
Goodfellow, I., Bengio, Y., Courville, A.: Deep Learning. MIT Press (2016),
  \url{http://www.deeplearningbook.org}

\bibitem{graves2016hybrid}
Graves, A., Wayne, G., Reynolds, M., Harley, T., Danihelka, I.,
  Grabska-Barwi{\'n}ska, A., Colmenarejo, S.G., Grefenstette, E., Ramalho, T.,
  Agapiou, J., et~al.: Hybrid computing using a neural network with dynamic
  external memory. Nature  \textbf{538}(7626), ~471 (2016)

\bibitem{hazan2010direct}
Hazan, T., Keshet, J., McAllester, D.A.: Direct loss minimization for
  structured prediction. In: Advances in Neural Information Processing Systems.
  pp. 1594--1602 (2010)

\bibitem{Keshavarz11}
Keshavarz, A., Wang, Y., Boyd, S.: Imputing a convex objective function. In:
  2011 {IEEE} International Symposium on Intelligent Control. pp. 613--619.
  IEEE (2011)

\bibitem{lecun2015deep}
LeCun, Y., Bengio, Y., Hinton, G.: Deep learning. Nature  \textbf{521},
  436--444 (2015)

\bibitem{Maclaurin15}
Maclaurin, D., Duvenaud, D., Adams, R.: Gradient-based hyperparameter
  optimization through reversible learning. In: International Conference on
  Machine Learning. pp. 2113--2122 (2015)

\bibitem{maddison2016concrete}
Maddison, C.J., Mnih, A., Teh, Y.W.: The concrete distribution: A continuous
  relaxation of discrete random variables. arXiv preprint arXiv:1611.00712
  (2016)

\bibitem{Mahmood18b}
Mahmood, R., Babier, A., McNiven, A., Diamant, A., Chan, T.C.Y.: Automated
  treatment planning in radiation therapy using generative adversarial
  networks. arXiv preprint arXiv:1807.06489  (2018)

\bibitem{pascanu2012understanding}
Pascanu, R., Mikolov, T., Bengio, Y.: Understanding the exploding gradient
  problem. CoRR, abs/1211.5063  (2012)

\bibitem{Paszke17}
Paszke, A., Gross, S., Chintala, S., Chanan, G., Yang, E., DeVito, Z., Lin, Z.,
  Desmaison, A., Antiga, L., Lerer, A.: Automatic differentiation in {PyTorch}.
  In: NeurIPS AutoDiff Workshop (2017)

\bibitem{convexHull}
{QHull Library}: {} (2018),
  \url{https://docs.scipy.org/doc/scipy-0.19.0/reference/generated/scipy.spatial.ConvexHull.html},
  [Online; accessed November 2, 2018]

\bibitem{Rumelhart86}
Rumelhart, D.E., Hinton, G.E., Williams, R.J.: Learning representations by
  back-propagating errors. Nature  \textbf{323}(6088),  533--536 (1986)

\bibitem{Gallego17}
Saez-Gallego, J., Morales, J.M.: Short-term forecasting of price-responsive
  loads using inverse optimization. IEEE Transactions on Smart Grid  (2017)

\bibitem{sahoo2018learning}
Sahoo, S.S., Lampert, C.H., Martius, G.: Learning equations for extrapolation
  and control. In: International Conference on Machine Learning (ICML) (2018)

\bibitem{soudry2017exponentially}
Soudry, D., Hoffer, E.: Exponentially vanishing sub-optimal local minima in
  multilayer neural networks. arXiv preprint arXiv:1702.05777  (2017)

\bibitem{sutskever2013training}
Sutskever, I.: Training recurrent neural networks. University of Toronto
  Toronto, Ontario, Canada (2013)

\bibitem{sutskever2013importance}
Sutskever, I., Martens, J., Dahl, G., Hinton, G.: On the importance of
  initialization and momentum in deep learning. In: International Conference on
  Machine Learning. pp. 1139--1147 (2013)

\bibitem{Troutt08}
Troutt, M.D., Brandyberry, A.A., Sohn, C., Tadisina, S.K.: Linear programming
  system identification: The general nonnegative parameters case. European
  Journal of Operational Research  \textbf{185}(1),  63--75 (2008)

\bibitem{Troutt05}
Troutt, M.D., Tadisina, S.K., Sohn, C., Brandyberry, A.A.: Linear programming
  system identification. European Journal of Operational Research
  \textbf{161}(3),  663--672 (2005)

\end{thebibliography}

\end{document}